\documentclass[a4paper,fleqn]{cas-dc}

\usepackage[authoryear,sort&compress]{natbib}

\usepackage{amssymb}
\usepackage{amsmath}
\usepackage{booktabs}
\usepackage{multirow}
\usepackage{array}
\usepackage{xcolor}
\usepackage{adjustbox}
\usepackage{placeins}

\begin{document}

\shorttitle{Externally Validated Breast Ultrasound Segmentation}
\shortauthors{J. Zhang et~al.}

\title[mode=title]{Externally Validated Breast Ultrasound Segmentation via Multi-task Learning with BI-RADS-Consistent Morphological Priors}

\author[1]{Jingru Zhang}

\author[2,5]{Saed Moradi}

\author[2,3,4]{Ashirbani Saha} [orcid=0000-0002-7650-1720]
\cormark[1]
\ead{sahaa16@mcmaster.ca}

\cortext[cor1]{Corresponding author.}
\affiliation[1]{organization={School of Biomedical Engineering, McMaster University},
            city={Hamilton},
            state={Ontario},
            country={Canada}}
\affiliation[2]{organization={Department of Oncology, McMaster University},
            city={Hamilton},
            state={Ontario},
            country={Canada}}
\affiliation[3]{organization={Centre for Data Science and Digital Health (CREATE), Hamilton Health Sciences},
            city={Hamilton},
            state={Ontario},
            country={Canada}}
\affiliation[4]{organization={Escarpment Cancer Research Institute (ECRI), McMaster University and Hamilton Health Sciences},
            city={Hamilton},
            state={Ontario},
            country={Canada}}
\affiliation[5]{organization={Vision and Image Processing Laboratory, Department of Systems Design Engineering, University of Waterloo},
            city={Waterloo},
            state={Ontario},
            country={Canada}}

\begin{abstract}
External validation of breast ultrasound segmentation models remains limited because internal train--test splits do not capture domain shifts across imaging systems, acquisition protocols, and patient populations. We introduce a novel multi-task framework for breast ultrasound segmentation and malignancy classification. Its central methodological advance is a differentiable morphology-to-malignancy bridge: lesion area, boundary roughness, compactness, and texture are computed from the predicted soft segmentation mask, aggregated with learned weights into a morphology-based malignancy score, and constrained to agree with the image-level classifier. To our knowledge, this is the first breast ultrasound framework to use BI-RADS-inspired morphology derived from its own soft segmentation output as an end-to-end consistency target for malignancy classification. It is also the first 2D B-mode multi-task study to report every directed external transfer among four independent datasets: training on each dataset and testing on the other three yields 12 source--target pairs, assessed with single models and five-fold ensembles. In matched comparisons across all pairs, the proposed single-model configuration outperforms dedicated single-task baselines in segmentation (mean Dice coefficient [DC]: 0.764 vs.\ 0.740) and malignancy classification (mean area under the receiver operating characteristic curve [AUC]: 0.818 vs.\ 0.791). The ensemble achieves a mean external DC of 0.786 and is competitive with SAM-based segmentation methods using an EfficientNet-B7 encoder while also predicting malignancy. The learned morphology weights retain the same ordering across all four datasets, with boundary roughness receiving the greatest weight. These results establish the first complete four-dataset directed benchmark for joint 2D breast ultrasound segmentation and malignancy classification and demonstrate that clinically grounded morphological consistency improves both tasks under domain shift.
\end{abstract}




\maketitle

\section{Introduction}
\label{sec:intro}

Breast cancer remains a leading cause of mortality among women worldwide, with early detection being critical for improving patient outcomes~\citep{sung2021global}. Breast ultrasound complements screening mammography, particularly for dense breast tissue~\citep{hooley2017screening}. Computer-aided diagnosis (CAD) systems for tumor segmentation and malignancy classification in ultrasound can assist radiologists~\citep{cheng2010automated}.

Deep learning has driven substantial progress in breast ultrasound segmentation and classification~\citep{aumente2025multi,he2024multi,zhang2021shamtl,chowdary2022multitask,lu2025mtloca,yang2025nmtnet,zhang2025uwtnet,hu2025unisegdiff,tang2025llm4seg,bruno2025dual,jiang2026sgla,wei2025nora,lin2025samus,zhuang2021abus}. However, the vast majority of these methods are developed and validated using internal train--test splits from a single center~\citep{zhuang2019rdaunet,almajalid2018development,vakanski2020attention,shareef2020stan,abraham2019focaltversky,xue2021global,huang2022boundary,tang2021fpnl,ning2022smunet,chen2022aaunet,chen2023rethinking,podda2022fully,iqbal2022mdanet,qi2023mdfnet,lyu2023amspan,roy2024daunet}. This evaluation design does not reflect the domain shift encountered when models are deployed across different ultrasound systems, imaging protocols, and patient populations~\citep{guan2022domain}. External evidence is now emerging, but it remains fragmented: segmentation studies commonly report selected source--target pairs~\citep{jiang2026sgla,wei2025nora,lin2025samus}; a joint Transformer reported diagnosis on one external cohort~\citep{zhang2024multitasktransformer}; joint 2D frameworks have been evaluated from one source on one or two external datasets~\citep{rahman2025hyformer,raheem2025fame}; and a 3D automated breast volume scanning (ABVS) model evaluated bidirectional transfer between one public and one private cohort~\citep{he2026bamt}. Systematic multi-dataset studies address segmentation alone~\citep{schmidt2025pca} or classification alone~\citep{wang2026shift}. Consequently, a complete directed external evaluation that measures both tasks and their interaction in one 2D B-mode framework is still missing. This is the gap we address; we review the relevant literature in detail in Section~\ref{sec:related}.

To address this gap, we propose a method designed to improve generalization: we adapt an EfficientNet-based multi-task design with a BI-RADS-inspired morphological prior and evaluate it under cross-dataset validation. Prior breast ultrasound models have used morphological and BI-RADS descriptors both for explainable malignancy diagnosis, as in BI-RADS-Net~\citep{zhang2021biradsnet}, and within multi-task designs~\citep{he2024multi}. We use BI-RADS-inspired descriptors differently: not as prediction targets or post-hoc explanations, but as a differentiable consistency signal computed from the predicted mask and evaluated across datasets. We propose a consistency regularizer that uses morphological features inspired by BI-RADS clinical descriptors~\citep{spak2017birads} to bridge the segmentation and classification tasks. This design is intended to reduce destructive task interference, which occurs when gradients from different task heads conflict during training and degrade performance~\citep{zhao2018modulationmodulemultitasklearning}. Because medical imaging datasets are inherently limited and expensive to annotate, the proposed strategy uses more of the available information: multi-task learning enables the segmentation model to benefit from image-level malignancy labels (benign/malignant); conversely, accurate segmentation provides morphological features that inform classification.

This study makes four distinct advances. \textbf{First,} we introduce a new task-coupling mechanism in which BI-RADS-inspired morphological features are computed differentiably from soft segmentation masks, aggregated through learned softmax weights into a composite malignancy score, and used as an explicit consistency target for the classification head. To our knowledge, no previous breast ultrasound framework has used morphology extracted from its own soft mask as an end-to-end bridge between segmentation and malignancy classification. This clinically grounded mechanism directly addresses destructive task interference rather than treating it only as a generic optimization problem. \textbf{Second,} we provide the first complete directed external evaluation of a 2D B-mode multi-task segmentation--classification model across four independent datasets: all 12 ordered train--test pairs spanning four countries, evaluated using both single models and five-fold ensembles. \textbf{Third,} within this same 12-pair protocol, we provide the first matched external comparison of clinically grounded task coupling against dedicated single-task networks, naive shared-encoder MTL, PCGrad, and uncertainty weighting. This design isolates whether the gain comes from the clinical relationship between tasks rather than from joint training or generic gradient management alone. \textbf{Fourth,} we identify a reproducible morphological hierarchy: the learned weights converge to the same ranking across all four datasets (Roughness $>$ Texture $>$ Area $>$ Compactness). This cross-domain stability provides interpretable evidence that morphology can serve as a robust link between lesion delineation and malignancy prediction.

\section{Related Work}
\label{sec:related}

\subsection{Breast ultrasound lesion segmentation}
The dominant paradigm for breast ultrasound lesion segmentation is the encoder--decoder architecture, with the U-Net family and its attention-augmented and multi-scale variants forming the backbone of most methods. Early adaptations modified U-Net for the speckle and low-contrast characteristics of ultrasound~\citep{almajalid2018development}, and subsequent work added residual and dilated attention gates to suppress background and sharpen boundaries (RDAU-Net~\citep{zhuang2019rdaunet}). A recurring theme is attention: salient-region attention enriched with task-specific priors~\citep{vakanski2020attention}, adaptive attention that learns lesion-generic representations (AAU-Net~\citep{chen2022aaunet}), dual attention along positional and windowed axes (DAU-Net~\citep{roy2024daunet}), and multiscale dual attention (MDA-Net~\citep{iqbal2022mdanet}). Others target the specific failure modes of breast ultrasound: small-tumor awareness (STAN~\citep{shareef2020stan}), saliency-guided and morphology-guided refinement (SMU-Net~\citep{ning2022smunet}), class imbalance via the focal Tversky loss~\citep{abraham2019focaltversky}, global context aggregation~\citep{xue2021global,tang2021fpnl}, explicit boundary rendering~\citep{huang2022boundary}, and multi-scale dynamic fusion (MDF-Net~\citep{qi2023mdfnet}, AMS-PAN~\citep{lyu2023amspan}, MEF-UNet~\citep{xu2024mefunet}). Recent work argues that carefully configured plain U-Nets remain highly competitive~\citep{chen2023rethinking}, while end-to-end pipelines couple segmentation with downstream classification~\citep{podda2022fully}. Beyond convolutional designs, very recent methods explore wavelet-domain feature mining~\citep{zhang2025uwtnet}, staged diffusion models~\citep{hu2025unisegdiff}, and semantic boosting using a pretrained large language model (LLM)~\citep{tang2025llm4seg}. Despite this methodological diversity, these works share a common evaluation protocol: development and testing on internal, single-center splits of one (or occasionally two) datasets, which leaves their behavior under inter-institutional domain shift uncharacterized. Section~\ref{sec:experiments} quantifies this point by comparing our internal performance against representative methods (Table~\ref{tab:internal_compare}) and then measuring what is lost under external validation.

\subsection{Multi-task learning for joint segmentation and classification}
Because breast ultrasound datasets typically carry both pixel-level masks and image-level malignancy labels, multi-task learning (MTL) that jointly performs segmentation and classification is attractive: each task supplies complementary supervision, and the shared encoder can learn richer representations than either task alone~\citep{aumente2025multi,he2024multi,zhang2021shamtl,chowdary2022multitask,lu2025mtloca,yang2025nmtnet}. \citet{aumente2025multi} report consistent gains of joint training over single-task baselines, and attention-based MTL designs (SHA-MTL~\citep{zhang2021shamtl}) further couple the two heads.

External testing of breast ultrasound MTL has precedents and is therefore not, by itself, the novelty of this study. \citet{zhang2024multitasktransformer} report diagnostic performance on a large external validation cohort for a joint segmentation--classification Transformer. \citet{bruno2025dual} include a shared-encoder multi-task Swin baseline in one cross-domain scenario, and the HyFormer-Net preprint examines zero-shot transfer and target-domain fine-tuning in one direction~\citep{rahman2025hyformer}. FAME trains on BUSI and reports both segmentation and classification on two independent test datasets, although its segmentation and classification components form a dual-stage federated ensemble rather than our shared end-to-end consistency design~\citep{raheem2025fame}. In 3D imaging, BAMT-Net jointly segments and classifies ABVS volumes and reports public-to-private and private-to-public zero-shot transfer~\citep{he2026bamt}. These studies establish that joint systems can be tested externally, but none reports every ordered transfer among four independent 2D B-mode datasets or evaluates a differentiable BI-RADS morphology consistency bridge.

The central difficulty of MTL is \emph{destructive task interference}: gradients from different heads can conflict on the shared parameters and degrade one or both tasks~\citep{zhao2018modulationmodulemultitasklearning}. General-purpose remedies operate on the optimization itself: PCGrad projects conflicting gradients onto orthogonal directions~\citep{yu2020pcgrad}, and uncertainty weighting learns per-task loss weights~\citep{kendall2018uncertainty}. These methods reduce interference but do so in a task-agnostic way, without exploiting the clinical relationship between a lesion's shape and its malignancy. Our approach instead introduces a domain-specific bridge: a consistency term that ties the classification head to morphological features computed from the segmentation output, so that the two tasks are encouraged to agree on clinically meaningful grounds rather than merely de-conflicted.

\subsection{Morphological and BI-RADS descriptors in breast ultrasound CAD}
Morphological descriptors of lesion shape and echotexture (area, margin irregularity, circularity/compactness, and internal echo heterogeneity) have underpinned breast ultrasound CAD for decades, and are formalized clinically in the BI-RADS lexicon~\citep{spak2017birads}, in which irregular margins and heterogeneous texture are among the strongest predictors of malignancy. Classical CAD systems computed such features from hand-segmented or thresholded masks and fed them to a separate classifier, breaking the pipeline into non-differentiable stages. More recent deep models reintroduce shape priors implicitly through attention or boundary modules~\citep{aslam2025hanet,he2024multi}. Explicit priors can also strengthen weak supervision: \citet{wang2026espc} incorporate an elliptical shape constraint into multiple-instance learning to improve ultrasound segmentation from coarse annotations. Their ellipse guides segmentation geometry and candidate-region construction for approximately elliptical targets. BAMT-Net fuses learned boundary semantics into a 3D multi-task network~\citep{he2026bamt}, but it does not derive multiple BI-RADS-inspired descriptors from the predicted soft mask or require a resulting morphology score to agree with the classifier. Our approach addresses a different problem and introduces a different role for prior knowledge: multiple BI-RADS-inspired descriptors are computed from the model's soft mask and become a trainable communication channel between segmentation and malignancy classification. Existing methods that adopt BI-RADS explicitly instead use it as a prediction target or post-hoc explanation. BI-RADS-Net~\citep{zhang2021biradsnet} predicts BI-RADS descriptors as auxiliary outputs that explain its malignancy prediction, and the segmentation-based BI-RADS ensemble classifier of \citet{bobowicz2024birads} uses BI-RADS categories as an ensemble target. In contrast, our consistency regularizer places the descriptors inside the end-to-end gradient path and requires the two task heads to agree through them. This differentiable morphology bridge, together with its complete directed external evaluation, constitutes the methodological and empirical novelty of our work.

\subsection{External validation and domain generalization}
Domain shift (e.g., differences in scanners, acquisition protocols, and patient populations across institutions) is a well-documented obstacle to deploying medical imaging models and has motivated extensive research on domain adaptation~\citep{guan2022domain}. Systematic cross-dataset evidence has recently appeared for individual breast ultrasound tasks. \citet{schmidt2025pca} examine segmentation across six datasets, whereas \citet{wang2026shift} provide a patient-leakage-aware pairwise benchmark for malignancy classification across four public datasets. Neither study jointly models or evaluates both tasks. External joint-system studies remain narrower: they use one training source and one or two external targets in 2D B-mode imaging~\citep{zhang2024multitasktransformer,bruno2025dual,rahman2025hyformer,raheem2025fame}, or two directions between one public and one private cohort in 3D ABVS~\citep{he2026bamt}.

A promising recent line of work adapts the Segment Anything Model (SAM) for domain-generalized ultrasound segmentation. SAMUS~\citep{lin2025samus} couples SAM with a CNN branch and reaches a Dice coefficient (DC) of 0.781 on BUSI$\rightarrow$UDIAT~\citep{yap2018automated} (we write $X\rightarrow Y$ for training on dataset $X$ and testing on dataset $Y$). Nora~\citep{wei2025nora} attains a DC of 0.839 through noise-robust SAM tuning, while SAMLR~\citep{jiang2026sgla} applies low-rank adaptation. These methods are effective but incur the cost of fine-tuning a foundation model. Because they perform segmentation only, they do not provide malignancy classification. The dual-stage framework of \citet{bruno2025dual} reports a DC of 0.738 on BUSI$\rightarrow$BrEaST~\citep{pawlowska2024breastlesionsusg}. Against this backdrop, our work provides the first complete single-source directed evaluation of a joint 2D B-mode model over all 12 train--test directions across four countries, rather than a selected pair or cohort. We also obtain competitive external performance with a conventional EfficientNet-B7 encoder, without foundation-model fine-tuning, while additionally producing a malignancy prediction.

\section{Proposed Method}
\label{sec:methods}

\begin{figure*}[t]
\centering
\includegraphics[width=\textwidth]{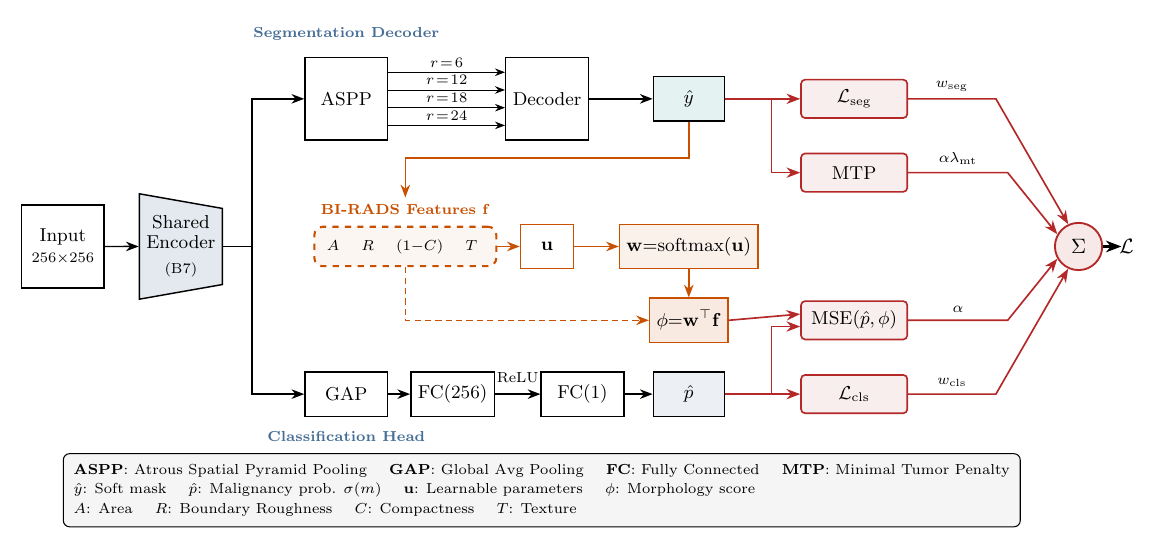}
\caption{Architecture overview. The shared EfficientNet-B7 encoder feeds a segmentation decoder (producing soft mask $\hat{y}$) and a classification head (producing malignancy probability $\hat{p}$). Our contribution (dashed orange): BI-RADS morphological features $\mathbf{f}$ computed from $\hat{y}$ are combined via learned softmax weights $\mathbf{w}$ into a morphology score $\phi$. The total loss $\mathcal{L}$ (Eq.~\ref{eq:loss}) aggregates all four terms.}
\label{fig:architecture}
\end{figure*}

Our method employs an encoder-decoder architecture (Figure~\ref{fig:architecture}) with a shared backbone for simultaneous segmentation and malignancy classification. The key innovation is a consistency regularization mechanism using BI-RADS-consistent morphological priors. We describe the model components below.

\textbf{Network Architecture.} We use an EfficientNet-B7~\citep{tan2019efficientnet} backbone pretrained on ImageNet as the encoder, with a DeepLabV3+~\citep{chen2018deeplab} decoder incorporating Atrous Spatial Pyramid Pooling (ASPP)~\citep{chen2017deeplab} for multi-scale feature extraction. The network produces two outputs: segmentation logits $s$ yielding soft masks $\hat{y} = \sigma(s)$, and malignancy logit $m$ yielding probability $\hat{p} = \sigma(m)$.

\textbf{BI-RADS Features.} We compute four morphological features from soft segmentation masks $\hat{y} = \sigma(s)$ as shown in Table~\ref{tab:features}. These features are inspired by BI-RADS clinical descriptors~\citep{spak2017birads} and can be optimized end-to-end. Boundary roughness and texture are normalized to $[0,1]$ using Exponential Moving Average (EMA) running statistics~\citep{ioffe2015batchnorm} for min-max normalization.

\begin{table*}[t]
\centering
\caption{BI-RADS-consistent morphological features computed from soft segmentation masks $\hat{y} = \sigma(s)$. All features are differentiable and can be optimized end-to-end. $R$ and $T$ are normalized to $[0,1]$ using exponential moving average (EMA) min-max statistics (momentum 0.99) computed over training batches.}
\label{tab:features}
\setlength{\tabcolsep}{3pt}
\fontsize{9}{11}\selectfont
\begin{adjustbox}{width=\textwidth}
\begin{tabular}{@{}p{0.13\textwidth} p{0.20\textwidth} p{0.53\textwidth}@{}}
\toprule
\textbf{Feature} & \textbf{Interpretation} & \textbf{Formula \& Computation} \\
\midrule
Area ($A$) & Larger masses: higher risk & $A = \text{mean}(\hat{y})$ computed over all pixels. \\[3pt]
Boundary Roughness ($R$) & Irregular margins: malignant & Sobel edge magnitude; perimeter $P{=}\sum E_{uv}$; roughness $R{=}P / \sqrt{\sum \hat{y}_{uv}}$ normalized to $[0,1]$ via EMA. \\[3pt]
Compactness ($C$) & Round (high $C$): benign & $C = \text{clip}(4\pi A / (P^2{+}\varepsilon),\, 0,\, 1)$ where $A{=}\sum \hat{y}_{uv}$, $P$ is perimeter, $\varepsilon{=}10^{-6}$. \\[3pt]
Texture ($T$) & Heterogeneous echo: malignant & $T = \text{var}_{\hat{y}}(g)$: mask-weighted variance of grayscale values $g$ within tumor region (Eq.~\ref{eq:texture}). Normalized via EMA. \\
\bottomrule
\end{tabular}
\end{adjustbox}
\end{table*}

The texture feature uses a mask-weighted variance to capture echo heterogeneity inside the predicted tumor region:
\begin{equation}
\label{eq:texture}
T = \frac{\sum_{u,v} \hat{y}_{uv}\,(g_{uv} - \bar{g}_{\hat{y}})^2}{\sum_{u,v} \hat{y}_{uv}}, \quad \bar{g}_{\hat{y}} = \frac{\sum_{u,v} \hat{y}_{uv}\,g_{uv}}{\sum_{u,v} \hat{y}_{uv}}
\end{equation}
where $T$ is the mask-weighted variance of grayscale intensity within the predicted tumor region, $g_{uv}$ is the grayscale pixel intensity, and $\hat{y}_{uv}$ is the soft mask probability at pixel $(u,v)$. Using the soft mask as weights keeps the operation differentiable with respect to the segmentation logits $s$.

\textbf{Composite Malignancy Prior.} We compute a morphological malignancy score $\phi$ by combining the four BI-RADS-consistent features (Area $A$, Boundary Roughness $R$, Compactness $C$, Texture $T$) through learned softmax weights: $\phi = \mathbf{w}^\top \mathbf{f},\; \phi\in[0,1]$, where $\mathbf{f} = [A, R, (1-C), T]^\top$ and $\mathbf{w} = [w_A, w_R, w_C, w_T]^\top$. The weight vector $\mathbf{w}$ is obtained via softmax normalization of learnable parameters $\mathbf{u}$, i.e., $\mathbf{w}=\text{softmax}(\mathbf{u})$. The weights are initialized uniformly, so all four features begin with equal weight (0.25); the weights are then learned via gradient descent.

\textbf{Training Objective.} The total loss combines task-specific and consistency terms:
\begin{equation}
\label{eq:loss}
\mathcal{L} = w_{\text{seg}}\mathcal{L}_{\text{seg}} + w_{\text{cls}}\mathcal{L}_{\text{cls}} + \alpha \cdot \text{MSE}(\hat{p}, \phi) + \alpha\lambda_{\text{mt}}\text{MTP}
\end{equation}
where $w_{\text{seg}}$ and $w_{\text{cls}}$ are task weights for segmentation and classification, respectively, $\alpha$ controls consistency strength, and $\lambda_{\text{mt}}$ weights the Minimal Tumor Penalty (MTP). The segmentation loss $\mathcal{L}_{\text{seg}}$ combines Dice loss and binary cross-entropy (BCE), while the classification loss $\mathcal{L}_{\text{cls}}$ uses BCE; the BCE components use the BCE-with-logits formulation. The consistency term $\text{MSE}(\hat{p}, \phi)$ penalizes disagreement between predicted malignancy $\hat{p}=\sigma(m)$ and morphology score $\phi$. The MTP is defined as $\text{MTP} = \frac{1}{N}\sum_{i} \mathbb{1}[\bar{y}_i < \tau]\,(\tau - \bar{y}_i)$, where $\mathbb{1}[\cdot]$ is the indicator function (equivalently, a Heaviside step on the stated condition), equal to 1 when $\bar{y}_i < \tau$ and 0 otherwise; $\bar{y}_i = \text{mean}(\hat{y}_i)$ is the mean predicted mask value for sample $i$; and $\tau{=}0.01$ is the threshold. The penalty activates only for samples whose mean mask prediction falls below $\tau$, in which case it grows as the prediction approaches zero, selectively penalizing near-empty predictions and preventing the segmenter from producing ambiguous low-confidence masks.

At inference, only the soft segmentation mask $\hat{y}$ and classification probability $\hat{p}$ are used; the morphological score $\phi$ and EMA statistics are not computed. This consistency loss is the key mechanism for mitigating destructive task interference: it creates an explicit bridge between segmentation and classification, encouraging the model to produce segmentation masks whose morphological characteristics align with the predicted malignancy assessment. Standard multi-task learning relies solely on shared representations; by contrast, this constraint enables beneficial task synergy.

\textbf{Training Configuration.} We train with the AdamW optimizer~\citep{loshchilov2019decoupled} using a learning rate of $6\times10^{-4}$ and weight decay $10^{-4}$. The effective batch size is 64 (batch size 16 with four gradient accumulation steps), with cosine annealing for up to 300 epochs and early stopping (patience 50). The extended training budget is motivated by the observation that overparameterized networks require substantially more optimization steps to generalize on small datasets~\citep{power2022grokking}. Task weights are $(w_{\text{seg}}, w_{\text{cls}}) = (0.9, 0.1)$. The consistency strength $\alpha$ was tuned per dataset (0.18 for BUSI, BrEaST, and BUS-BRA~\citep{gomez2023busbra}; 0.25 for UDIAT). The MTP weight $\lambda_{\text{mt}}$ was set to 1.0. All images are resized to $256\times256$.

\textbf{Data Augmentation.} We employ offline chain augmentation~\citep{cubuk2020randaugment,chen2022advchain} to expand each dataset to 5,000 samples. Related approaches include RandAugment~\citep{cubuk2020randaugment}, which uses uniform magnitude selection, and AdvChain~\citep{chen2022advchain}, which optimizes transformation parameters adversarially for medical image segmentation. Our approach randomly chains transformations to reach fixed dataset-specific target sample counts. For each sample, we apply one to three geometric transformations, zero to two intensity transformations, and zero or one noise transformation. Geometric transformations are selected uniformly from a pool of 11 operations: horizontal and vertical flips; rotations at $\pm15^{\circ}$, $\pm30^{\circ}$, and $\pm45^{\circ}$; shift-scale with limits 0.1/0.15; elastic deformation with $\alpha{=}50$ and $\sigma{=}2.5$; grid and optical distortion; perspective warp; and random resized cropping at scale 0.7--1.0. Intensity transformations include brightness/contrast adjustment ($\pm0.2$), gamma correction (80--120), CLAHE~\citep{zuiderveld1994clahe} with clipping limits of 1--4, sharpening, and equalization. Noise transformations include Gaussian noise ($\sigma^2{=}5$--25), Gaussian or median blurring ($k{=}3$--5), and multiplicative speckle noise (0.9--1.1). Geometric transformations are applied jointly to the image and mask; intensity and noise transformations are applied only to the image. Chain augmentation is applied only after the train/validation/test split, so no augmented version of any validation or test image is ever seen during training.

\section{Datasets and Experimental Setup}
\label{sec:datasets}

We evaluate on four publicly available breast ultrasound datasets from different countries (Table~\ref{tab:datasets}). The datasets range in size from 163 to 1,875 images and represent diverse ultrasound equipment and acquisition protocols. The \textbf{BUSI} dataset~\citep{aldhabyani2020busi} from Egypt contains 647 images (437 benign, 210 malignant) from 600 female patients aged 25--75. The \textbf{BrEaST} dataset~\citep{pawlowska2024breastlesionsusg} from Poland comprises 252 images collected across five medical centers using various ultrasound systems; all cases were confirmed by biopsy or follow-up. The \textbf{UDIAT} dataset~\citep{yap2018automated} from Spain includes 163 images (110 benign, 53 malignant). The \textbf{BUS-BRA} dataset~\citep{gomez2023busbra} from Brazil provides 1,875 biopsy-proven images from 1,064 patients acquired using four ultrasound systems at a single institution. These datasets exhibit considerable heterogeneity that poses challenges for cross-dataset generalization: resolution varies by more than 2$\times$, tumor size relative to image area ranges from 4.8\% to 9.3\%, and equipment diversity ranges from single-system acquisitions to multi-center collection.

\textbf{Evaluation Protocols.} We employ two complementary protocols for assessing cross-dataset generalization. \textit{Five-fold ensemble evaluation.} We perform five-fold stratified cross-validation with patient-level splits (80\% train, 20\% test per fold) using \texttt{StratifiedGroupKFold} (scikit-learn). Per-fold leakage checks verify that no patient appears in both the training and test folds. Within each fold, the training portion is further split for hyperparameter selection and overfitting monitoring. For internal validation, each fold-specific model is evaluated on its held-out 20\% test fold. Chain augmentation, when used, is applied only to the training portion of each split; no augmented version of any validation or test image is seen during training. For external evaluation, all five fold-specific models jointly predict on each external dataset. The final segmentation mask is obtained by pixel-wise majority vote across the five binary masks. The area under the receiver operating characteristic curve (AUC) for classification is computed from the mean predicted malignancy probability across the five models; hard malignancy labels, where reported, use sample-wise majority voting. This \emph{ensemble approach} uses all five trained models, maximizing the utility of the cross-validation procedure. \textit{Single-model evaluation.} Using the best hyperparameters identified from the five-fold cross-validation, we train a single model on each full dataset, typically with chain augmentation, using an 85/15 patient-level train/validation split. The validation portion is used only for early stopping.

With four datasets, each single model or ensemble is evaluated on the three remaining datasets, yielding 12 cross-dataset pairs per model type. The next section reports both internal-validation and external cross-dataset performance.

\begin{table*}[t]
\centering
\caption{Dataset characteristics. B/M denotes the benign/malignant split. Tumor size represents the average ratio of tumor pixels to image pixels. All cross-validation and cross-dataset splits are performed at the patient level, so no patient contributes images to more than one fold or across the train--test boundary.}
\label{tab:datasets}
\fontsize{9}{11}\selectfont
\setlength{\tabcolsep}{6pt}
\begin{tabular}{@{}lcccc@{}}
\toprule
\textbf{Dataset (Country)} & \textbf{N (B/M)} & \textbf{Resolution} & \textbf{Tumor Size} & \textbf{Scanner} \\
\midrule
BUSI~\citep{aldhabyani2020busi} (Egypt) & 647 (437/210) & $608\times495$ & 9.3\% & LOGIQ E9 \\
BrEaST~\citep{pawlowska2024breastlesionsusg} (Poland) & 252 (154/98) & $627\times523$ & 7.2\% & Various \\
UDIAT~\citep{yap2018automated} (Spain) & 163 (110/53) & $538\times455$ & 4.8\% & Sequoia C512 \\
BUS-BRA~\citep{gomez2023busbra} (Brazil) & 1,875 (1268/607) & $321\times391$ & 8.7\% & 4 systems \\
\bottomrule
\end{tabular}
\end{table*}

\section{Results}
\label{sec:experiments}

\subsection{Internal Validation}

Table~\ref{tab:internal} presents the five-fold cross-validation evaluation results. Models internally validated on BUS-BRA achieve the highest segmentation DC (0.910), which is attributable to the dataset's large training size. The models achieve DC values above 0.80 on all four datasets. Models internally validated on BUSI achieve the highest classification AUC (0.963), while those internally validated on BrEaST have the lowest AUC (0.861).

\begin{table*}[t]
\centering
\caption{Internal validation using five-fold cross-validation with chain augmentation. DC, IoU, and HD95 are reported for segmentation; AUC and F1 are reported for classification. Values are the mean $\pm$ standard deviation across folds.}
\label{tab:internal}
\fontsize{9}{11}\selectfont
\begin{tabular}{@{}lccccc@{}}
\toprule
\textbf{Dataset} & \textbf{DC} & \textbf{IoU} & \textbf{HD95} & \textbf{AUC} & \textbf{F1} \\
\midrule
BUS-BRA & $.910{\pm}.007$ & $.835{\pm}.010$ & $15.1{\pm}2.8$ & $.950{\pm}.012$ & $.857{\pm}.026$ \\
UDIAT & $.895{\pm}.018$ & $.822{\pm}.022$ & $12.0{\pm}4.8$ & $.918{\pm}.028$ & $.746{\pm}.079$ \\
BrEaST & $.828{\pm}.029$ & $.718{\pm}.035$ & $18.5{\pm}4.0$ & $.861{\pm}.045$ & $.731{\pm}.038$ \\
BUSI & $.809{\pm}.015$ & $.690{\pm}.018$ & $28.5{\pm}5.5$ & $.963{\pm}.018$ & $.891{\pm}.028$ \\
\bottomrule
\end{tabular}
\end{table*}

\subsection{Cross-Dataset Evaluation}
\label{sec:crossdataset}

Tables~\ref{tab:cross_dice}, \ref{tab:cross_hd95}, and \ref{tab:cross_auc} present the cross-dataset results. Segmentation DC (Table~\ref{tab:cross_dice}), boundary distance HD95 (Table~\ref{tab:cross_hd95}), and classification AUC (Table~\ref{tab:cross_auc}) are reported for both single-model (S) and five-fold ensemble (E) evaluation. Four patterns emerge. First, ensemble voting consistently improves on single-model predictions; for example, the DC for BUSI$\rightarrow$BrEaST increases from 0.720 to 0.753 (+4.7\%). Second, the BrEaST dataset is the best-generalizing training source according to mean ensemble DC (0.816), whereas the largest dataset, BUS-BRA, yields a mean of 0.795. Thus, training-set size alone does not determine generalization. Third, cross-dataset generalization is asymmetric: the DC is 0.826 for BUSI$\rightarrow$UDIAT and 0.691 for the reverse direction. Fourth, classification AUC generalizes more consistently, exceeding 0.80 in 11 of 12 pairs under ensemble evaluation; only UDIAT$\rightarrow$BrEaST falls below this threshold, with an AUC of 0.738.

\begin{table*}[t]
\centering
\caption{Cross-dataset segmentation performance measured by DC (higher is better). S denotes a single model, and E denotes a five-fold ensemble using majority voting. Columns are external targets. The best value in each column is shown in \textbf{bold}. The bottom \textit{Target mean} row averages each target column over its three source datasets.}
\label{tab:cross_dice}
\fontsize{9}{11}\selectfont
\setlength{\tabcolsep}{6pt}
\begin{tabular}{@{}llcccccc@{}}
\toprule
\textbf{Train} & \textbf{Prot.} & \textbf{BU} & \textbf{Br} & \textbf{UD} & \textbf{BB} & \textbf{Avg} \\
\midrule
BUSI (BU)    & S & -- & .720 & .795 & .809 & .775 \\
             & E & -- & .753 & .826 & .837 & .806 \\
\addlinespace[1pt]
BrEaST (Br)  & S & .714 & -- & .839 & .824 & .792 \\
             & E & .734 & -- & \textbf{.865} & \textbf{.849} & \textbf{.816} \\
\addlinespace[1pt]
UDIAT (UD)   & S & .675 & .680 & -- & .773 & .709 \\
             & E & .691 & .693 & -- & .799 & .728 \\
\addlinespace[1pt]
BUS-BRA (BB) & S & .744 & .772 & .825 & -- & .780 \\
             & E & \textbf{.757} & \textbf{.786} & .842 & -- & .795 \\
\midrule
\textbf{Target mean} & S & .711 & .724 & .820 & .802 & .764 \\
             & E & .727 & .744 & .844 & .828 & .786 \\
\bottomrule
\end{tabular}
\end{table*}

\begin{table*}[t]
\centering
\caption{Cross-dataset segmentation boundary accuracy measured by the median 95th-percentile Hausdorff distance (HD95) in pixels on $256\times256$ images (lower is better). Results are shown for the single-model (S) and five-fold ensemble (E) configurations, matching the S/E layout of Tables~\ref{tab:cross_dice} and~\ref{tab:cross_auc}. We report the median rather than the mean because HD95 is heavy-tailed: a few total-miss cases produce very large distances that would dominate the mean. The lowest average for each configuration is shown in \textbf{bold}. The bottom \textit{Target mean} row averages each target column over its three source datasets.}
\label{tab:cross_hd95}
\fontsize{9}{11}\selectfont
\setlength{\tabcolsep}{5pt}
\begin{tabular}{@{}llccccc@{}}
\toprule
\textbf{Train} & & \textbf{BU} & \textbf{Br} & \textbf{UD} & \textbf{BB} & \textbf{Avg} \\
\midrule
BUSI (BU)    & S & -- & 12.08 & 4.47 & 10.36 & \textbf{8.97} \\
             & E & -- & 9.43 & 5.00 & 9.00 & \textbf{7.81} \\
BrEaST (Br)  & S & 25.57 & -- & 5.00 & 13.79 & 14.79 \\
             & E & 13.00 & -- & 4.24 & 8.00 & 8.41 \\
UDIAT (UD)   & S & 20.00 & 13.08 & -- & 11.70 & 14.93 \\
             & E & 16.01 & 11.95 & -- & 8.53 & 12.16 \\
BUS-BRA (BB) & S & 16.12 & 9.63 & 5.00 & -- & 10.25 \\
             & E & 13.89 & 8.06 & 4.60 & -- & 8.85 \\
\midrule
\textbf{Target mean} & S & 20.56 & 11.60 & 4.82 & 11.95 & 12.23 \\
             & E & 14.30 & 9.81 & 4.61 & 8.51 & 9.31 \\
\bottomrule
\end{tabular}
\end{table*}

\begin{table*}[t]
\centering
\caption{Cross-dataset classification performance measured by malignancy AUC (higher is better). S denotes a single model, and E denotes a five-fold ensemble. The best value in each column is shown in \textbf{bold}. The bottom \textit{Target mean} row averages each target column over its three source datasets.}
\label{tab:cross_auc}
\fontsize{9}{11}\selectfont
\setlength{\tabcolsep}{6pt}
\begin{tabular}{@{}llcccccc@{}}
\toprule
\textbf{Train} & \textbf{Prot.} & \textbf{BU} & \textbf{Br} & \textbf{UD} & \textbf{BB} & \textbf{Avg} \\
\midrule
BUSI (BU)    & S & -- & .782 & .887 & .802 & .824 \\
             & E & -- & .809 & .898 & .847 & .851 \\
\addlinespace[1pt]
BrEaST (Br)  & S & .866 & -- & .803 & .809 & .826 \\
             & E & \textbf{.898} & -- & .827 & \textbf{.853} & .859 \\
\addlinespace[1pt]
UDIAT (UD)   & S & .829 & .714 & -- & .775 & .773 \\
             & E & .885 & .738 & -- & .816 & .813 \\
\addlinespace[1pt]
BUS-BRA (BB) & S & .867 & .789 & .891 & -- & .849 \\
             & E & .892 & \textbf{.821} & \textbf{.913} & -- & \textbf{.875} \\
\midrule
\textbf{Target mean} & S & .854 & .762 & .860 & .795 & .818 \\
             & E & .892 & .789 & .879 & .839 & .850 \\
\bottomrule
\end{tabular}
\end{table*}

The source asymmetry summarized above is most pronounced between BUSI and UDIAT and appears to be driven by tumor size: BUSI's larger tumors (9.3\% of image area) provide features that transfer to UDIAT's smaller tumors (4.8\%) more readily than the reverse. To substantiate this tumor-size hypothesis, we stratified the per-image external predictions by lesion size, binning all 2,937 images pooled from the four datasets into terciles (small $[0,0.037)$, medium $[0.037,0.093)$, and large $[0.093,0.5754]$ of image area). The exact cut points were 0.0369 and 0.0925 at original resolution (0.0369 and 0.0926 after resizing), which round to the reported 0.037 and 0.093. Table~\ref{tab:size_strat} reports results for the two most asymmetric directions, BUSI$\rightarrow$UDIAT and its reverse. The two directions show opposite lesion-size effects. The BUSI-trained model, which was trained on larger tumors, is most accurate on large external lesions and performs less well on small lesions (DC $0.893\rightarrow0.861\rightarrow0.770$ from large to small). By contrast, the UDIAT-trained model, which was trained on smaller tumors, is most accurate on small external lesions and degrades substantially on large lesions (DC $0.746$ on small vs.\ $0.607$ on large). HD95 shows the same asymmetry more clearly: for UDIAT$\rightarrow$BUSI, the median HD95 rises monotonically from $3.6$ pixels for small lesions to $23.4$ pixels for medium lesions and $46.1$ pixels for large lesions. This pattern supports the tumor-size explanation for the observed asymmetry. For BUSI$\rightarrow$UDIAT, DC and HD95 both increase with lesion size: DC reaches 0.893 and median HD95 reaches 10.0\,px for large lesions. These results are not contradictory because HD95 is an absolute pixel distance; a large lesion can have greater overlap while retaining a larger absolute boundary error.


The learned morphological weights converge to a consistent ranking across all four datasets: Roughness (0.29--0.33), Texture (0.25--0.27), Area (0.23--0.26), and Compactness (0.15--0.23). The dominance of boundary roughness aligns with clinical evidence that irregular margins are among the strongest BI-RADS predictors of malignancy~\citep{spak2017birads}, and this consistency across international datasets suggests that the morphology-to-malignancy relationship is robust to domain shift or data drift arising from differences in data acquisition.

\begin{table*}[t]
\centering
\caption{Tumor-size-stratified external segmentation performance for the two most asymmetric cross-dataset directions, BUSI$\leftrightarrow$UDIAT. We chose BUSI and UDIAT because they lie at opposite ends of the tumor-size distribution among our four datasets (mean tumor area: BUSI, $9.3\%$; UDIAT, $4.8\%$; see Table~\ref{tab:datasets}) and exhibit the largest asymmetry in the cross-dataset matrix (ensemble DC: BUSI$\rightarrow$UDIAT, $0.826$; reverse direction, $0.691$; a $13.5$-point gap). Mean DC and median HD95 in pixels are reported for each pooled tumor-area tercile. We summarize HD95 by its median because its per-image distribution is heavy-tailed: total-miss predictions (DC $\approx$ 0) produce very large boundary distances that would dominate the mean, whereas the median reflects the typical error. DC is bounded in $[0,1]$ and is well summarized by its mean. Models trained on BUSI transfer best to large lesions, whereas models trained on UDIAT transfer best to small lesions.}
\label{tab:size_strat}
\fontsize{9}{11}\selectfont
\setlength{\tabcolsep}{6pt}
\begin{tabular}{@{}llccc@{}}
\toprule
\textbf{Direction} & \textbf{Metric} & \textbf{Small} & \textbf{Medium} & \textbf{Large} \\
\midrule
\multirow{2}{*}{BUSI $\rightarrow$ UDIAT (forward)} & DC $\uparrow$ & .770 & .861 & \textbf{.893} \\
                                                    & HD95 (med) $\downarrow$ & \textbf{3.16} & 4.79 & 10.00 \\
\addlinespace[1pt]
\multirow{2}{*}{UDIAT $\rightarrow$ BUSI (reverse)} & DC $\uparrow$ & \textbf{.746} & .609 & .607 \\
                                                    & HD95 (med) $\downarrow$ & \textbf{3.60} & 23.35 & 46.10 \\
\bottomrule
\end{tabular}
\end{table*}

\subsection{Comparison with Existing Methods}

To position our internal performance relative to the published literature, Table~\ref{tab:internal_compare} compares our five-fold cross-validation DC with representative breast ultrasound segmentation methods for which BUSI and/or UDIAT DC values are reported in the original papers. We deliberately restrict the table to values that we could verify against primary sources. Methods that report results on private datasets (e.g., RDAU-Net~\citep{zhuang2019rdaunet}, STAN~\citep{shareef2020stan}, and the attention-enriched method of \citet{vakanski2020attention}) or use preprocessing that we could not match (GG-Net~\citep{xue2021global}, MDF-Net~\citep{qi2023mdfnet}, and ``Rethinking U-Net''~\citep{chen2023rethinking}) are cited in Section~\ref{sec:related} but omitted here to avoid a misleading comparison. Two caveats apply even to the included rows: (i) each paper uses its own split protocol (four-fold cross-validation, 80/20, or k-fold with different patient handling) and preprocessing, so absolute values are not strictly comparable; and (ii) our model simultaneously predicts malignancy and is validated across centers (Section~\ref{sec:crossdataset}), whereas the segmentation-only entries provide neither capability. Three of the included methods (\citet{podda2022fully}, DBU-Net~\citep{pramanik2023dbunet}, and USE-MiT~\citep{brancati2026use}) report higher in-domain BUSI DC values than ours under their own protocols. However, USE-MiT's strong in-domain four-fold BUSI value (0.868) does not carry over to cross-dataset evaluation (Section~\ref{sec:crossdataset}, Table~\ref{tab:sota}), which is the more informative test. On BrEaST, the recent text-guided multimodal model XBusNet~\citep{mallina2025xbusnet} reports a DC of 0.877 (vs.\ 0.828 for our method) while additionally using BI-RADS text prompts at inference. Table~\ref{tab:internal_compare} is therefore intended only to provide representative context rather than a complete leaderboard. Because protocols, splits, and preprocessing differ substantially across studies, the cross-dataset experiments in Sections~\ref{sec:crossdataset}--\ref{sec:ablation} are the primary basis for evaluating generalization.

\begin{table*}[t]
\centering
\caption{Internal-validation comparison with representative single-center breast ultrasound segmentation methods, restricted to BUSI and UDIAT, the only datasets used by the included methods. Our five-fold patient-level cross-validation DC values for BrEaST (0.828) and BUS-BRA (0.910) are reported in Table~\ref{tab:internal}; existing studies did not evaluate these datasets. The first row reproduces our five-fold cross-validation mean from Table~\ref{tab:internal}. Competing values are reported as given in the original papers under their own splits and preprocessing and are \emph{not strictly comparable} (see text); ``--'' indicates that a method did not use the dataset. $^\ddagger$The value was reported in published benchmark comparisons under a common BUSI protocol~\citep{wang2025dcceunet} rather than in the original paper.}
\label{tab:internal_compare}
\fontsize{9}{11}\selectfont
\setlength{\tabcolsep}{6pt}
\begin{tabular}{@{}llcc@{}}
\toprule
\textbf{Method} & \textbf{Architecture} & \textbf{BUSI} & \textbf{UDIAT} \\
\midrule
\textbf{Ours} & \textbf{EfficientNet-B7 + DLv3+ (MT)} & .809 & .895 \\
\midrule
HEATNet$^\ddagger$~\citep{jiang2023heatnet} & Hybrid enhanced attention transformer & .741 & -- \\
DAU-Net~\citep{roy2024daunet}             & Dual-attention U-Net          & .742 & .786 \\
UNeXt$^\ddagger$~\citep{valanarasu2022unext} & MLP-based U-Net             & .747 & -- \\
LAEDNet$^\ddagger$~\citep{zhou2022laednet} & Lightweight attention encoder--decoder & .750 & -- \\
SC-UNeXt~\citep{cai2024scunext}           & Lightweight UNeXt + SCConv    & .753 & -- \\
AAU-Net~\citep{chen2022aaunet}            & Adaptive-attention U-Net      & .775 & .781 \\
CFU-Net$^\ddagger$~\citep{yin2023cfunet}  & Coarse--fine U-Net w/ multilevel attention & .786 & -- \\
SMU-Net~\citep{ning2022smunet}            & Saliency-/morphology-aware    & .794 & -- \\
DCCE-UNet~\citep{wang2025dcceunet}        & Difference-/context-aware U-Net & .799 & -- \\
\citet{podda2022fully}       & Automated seg.+cls.\ pipeline  & .826 & -- \\
DBU-Net~\citep{pramanik2023dbunet}        & Dual-branch U-Net (edge-enh.)  & .853$^\ddagger$ & .873 \\
USE-MiT~\citep{brancati2026use}           & MiT-B4 U-Net + SE attention    & .868 & -- \\
\bottomrule
\end{tabular}
\end{table*}

Table~\ref{tab:sota} compares our method with published cross-dataset segmentation results and with methods that we reimplemented or reproduced. Published \emph{per-method} external baselines exist for only two of the 12 cross-dataset pairs (BUSI$\rightarrow$UDIAT and BUSI$\rightarrow$BrEaST). Cross-dataset generalization in breast ultrasound has been examined more broadly only recently. \citet{schmidt2025pca} report a PCA-preprocessing study over six datasets, but their analysis addresses segmentation only, reporting pixel-level recall and DC without malignancy classification, and uses a different dataset pool and evaluation protocol. In addition, the study is a preprint and is not directly comparable to the results reported here. To our knowledge, no peer-reviewed study has yet reported a systematic directed cross-dataset benchmark spanning all train--test pairs among these four public datasets while jointly addressing malignancy classification and BI-RADS morphological interpretability. Where published external results are unavailable, direct comparison is not possible without reimplementation under a matched protocol. We therefore reimplemented HA-Net~\citep{aslam2025hanet} and USE-MiT~\citep{brancati2026use} and reproduced the publicly released SAM-based Nora model~\citep{wei2025nora}. We evaluated all three methods on the external target datasets using BUSI as the training source; we also evaluated the two reimplemented models across all 12 cross-dataset pairs.

\begin{table*}[t]
\centering
\caption{Cross-dataset segmentation comparison measured by DC. All methods are trained on BUSI and evaluated on external datasets. \emph{Published} values are taken from the original papers, each of which reports a single external target: UDIAT results are from \citet{jiang2026sgla}, \citet{wei2025nora}, and \citet{lin2025samus}; results on the BrEaST dataset marked $^\dagger$ are from \citet{bruno2025dual}. In the \textit{Reimplemented/reproduced} block (marked $^*$), we reimplemented HA-Net and USE-MiT and reproduced Nora. We evaluated these methods on all three external targets under our unified protocol ($256\times256$, identical metric and method-specific preprocessing); our reproduced Nora result on BUSI$\rightarrow$UDIAT (0.839) matches its published value. Our values reproduce the BUSI-source ensemble row of Table~\ref{tab:cross_dice}; they are pair-specific BUSI-source results rather than the target means in that table. The best value in each column is shown in \textbf{bold}.}
\label{tab:sota}
\fontsize{8}{10}\selectfont
\setlength{\tabcolsep}{5pt}
\begin{tabular}{@{}llccc@{}}
\toprule
\textbf{Method} & \textbf{Architecture} & \textbf{BrEaST} & \textbf{UDIAT} & \textbf{BUS-BRA} \\
\midrule
\multicolumn{5}{@{}l}{\textit{Published results}} \\
UNet~\citep{jiang2026sgla} & Encoder-Decoder & -- & .750 & -- \\
TransUNet~\citep{jiang2026sgla} & CNN + Transformer & -- & .720 & -- \\
SwinUNet~\citep{jiang2026sgla} & Swin Transformer & -- & .730 & -- \\
UNet++~\citep{jiang2026sgla} & Nested skip conn. & -- & .750 & -- \\
DeepLabv3+~\citep{jiang2026sgla} & ResNet backbone & -- & .780 & -- \\
SAMLR~\citep{jiang2026sgla} & SAM + LoRA & -- & .805 & -- \\
SAMUS~\citep{lin2025samus} & SAM + CNN & -- & .781 & -- \\
Mask R-CNN$^\dagger$~\citep{bruno2025dual} & ResNet-50 + FPN & .523 & -- & -- \\
Multi-task Transf.$^\dagger$~\citep{bruno2025dual} & ViT-based & .560 & -- & -- \\
Dual-Stage DL$^\dagger$~\citep{bruno2025dual} & DLv3+ResNet34 + MNv3 & .738 & -- & -- \\
\midrule
\multicolumn{5}{@{}l}{\textit{Reimplemented/reproduced (our unified protocol)}$^*$} \\
USE-MiT~\citep{brancati2026use} & SegFormer MiT-B4 & .725 & .785 & .790 \\
HA-Net~\citep{aslam2025hanet} & DenseNet-121 + Hybrid Att. & .623 & .676 & .636 \\
Nora~\citep{wei2025nora} & SAM ViT-B & \textbf{.759} & \textbf{.839} & .821 \\
\midrule
\multicolumn{5}{@{}l}{\textit{Ours}} \\
Ensemble model & EfficientNet-B7 + DLv3+ & .753 & .826 & \textbf{.837} \\
\bottomrule
\end{tabular}
\end{table*}

\begin{figure*}[t]
\centering
\includegraphics[width=0.97\textwidth]{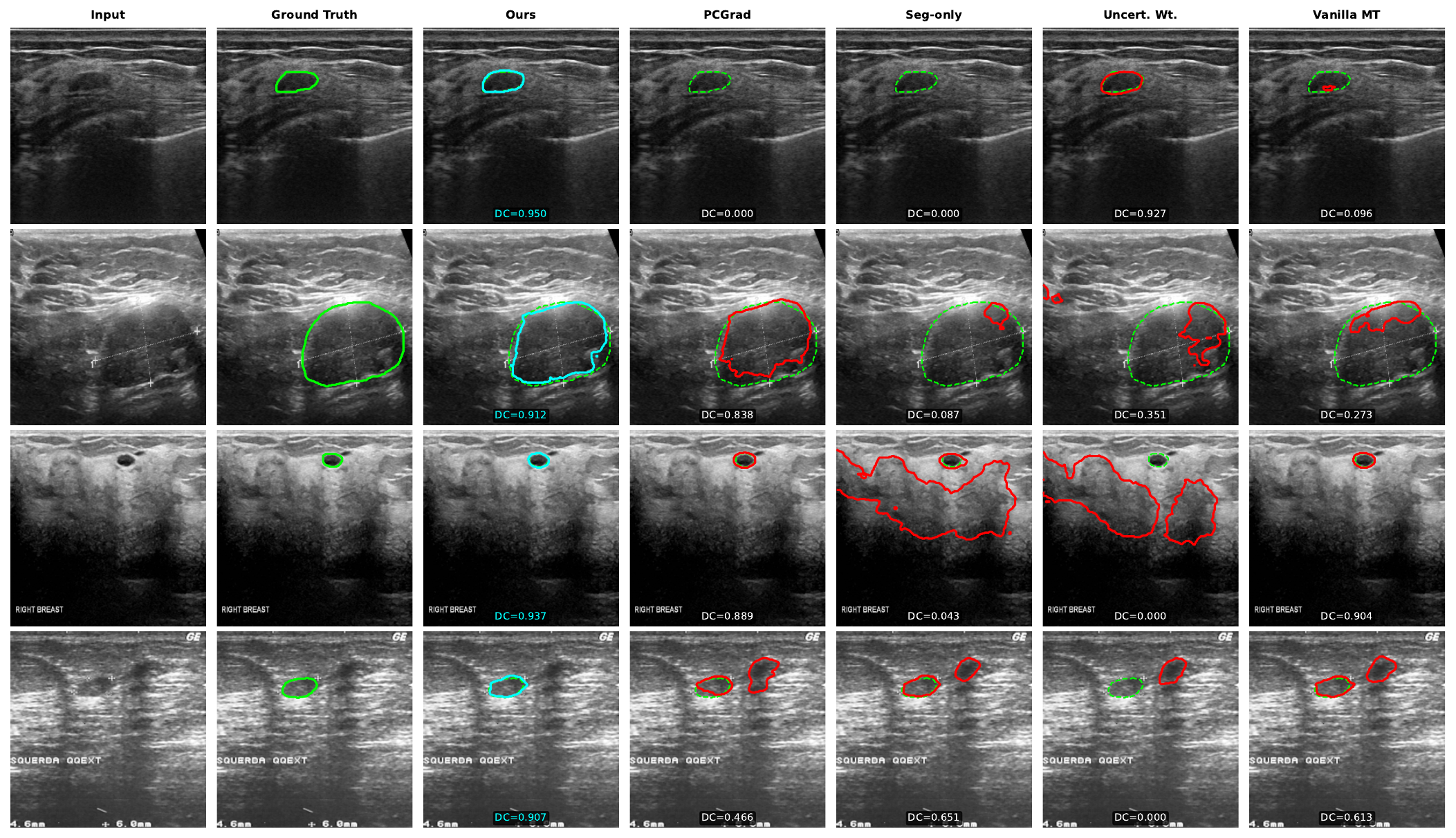}
\caption{Qualitative cross-dataset segmentation results comparing our method with baselines. Green contours in the prediction columns indicate the ground-truth boundaries.}
\label{fig:seg_comparison}
\end{figure*}

\begin{figure*}[!htbp]
\centering
\includegraphics[width=0.65\textwidth]{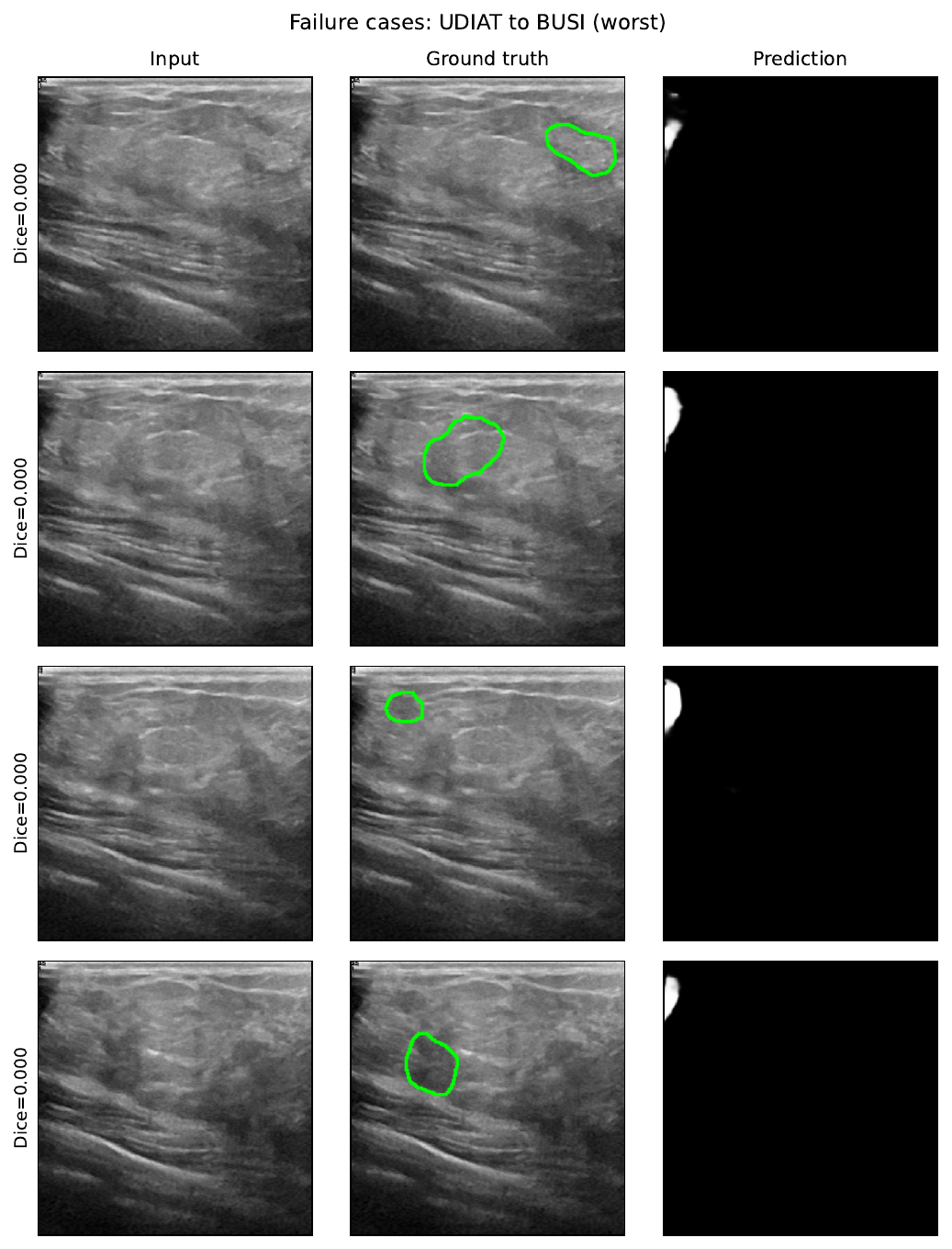}
\caption{Representative failure cases for the worst-performing direction, UDIAT$\rightarrow$BUSI (single-model DC 0.675; ensemble DC 0.691). Each row shows the input BUSI image, the ground-truth contour (green), and the model's soft prediction. The UDIAT-trained model misses every annotated tumor (DC $=$ 0.000 across all four cases) and instead consistently activates on a small region in the upper-left corner of each image. The activation appears to coincide with a dataset-specific annotation marker present in BUSI at that location but absent from UDIAT, suggesting that the model relies on a spurious cue that was absent from the UDIAT training data. This is a clear instance of dataset-specific spurious features rather than generic segmentation difficulty, and it accompanies the size-mismatch effect quantified in Table~\ref{tab:size_strat}.}
\label{fig:seg_failure}
\end{figure*}

On BUSI$\rightarrow$UDIAT, our ensemble model reaches a DC of 0.826. This value is close to that of the SAM-based Nora model~\citep{wei2025nora} (DC $=$ 0.839, both as published and as reproduced here) and exceeds those of SAMUS~\citep{lin2025samus} (0.781) and SAMLR~\citep{jiang2026sgla} (0.805). Our model uses a conventional EfficientNet-B7 encoder without foundation-model pretraining and also predicts malignancy, whereas the comparison methods perform segmentation only. These published figures were obtained under each method's own evaluation protocol rather than the unified directed cross-dataset protocol used here. USE-MiT illustrates the gap: it reports a DC of 0.88 on BUSI$\rightarrow$BrEaST under its own protocol (0.868 in-domain on BUSI), yet our faithful reimplementation under the unified protocol reaches only 0.725. Published values are therefore best interpreted as reference points rather than as a matched head-to-head comparison. On BUSI$\rightarrow$BrEaST, our ensemble model achieves a DC of 0.753, surpassing the Dual-Stage framework~\citep{bruno2025dual} (DC $=$ 0.738). Averaged over the three external target datasets evaluated with BUSI as the training source, our ensemble model (mean DC, 0.805) is on par with the reproduced Nora model (mean DC, 0.806), despite Nora's SAM foundation backbone and our method's additional malignancy prediction. Our method also clearly outperforms the reimplemented USE-MiT (0.767) and HA-Net (0.645). Across the full 12-pair matrix, our method attains a mean DC of 0.786 and exceeds the reimplemented USE-MiT (mean DC, 0.719; the full 12-pair result is not shown in Table~\ref{tab:sota}) on all 12 pairs. Figure~\ref{fig:seg_comparison} shows qualitative segmentation comparisons across methods, and Figure~\ref{fig:seg_failure} shows representative failure cases under domain shift.

\subsection{Ablation Study}
\label{sec:ablation}

Table~\ref{tab:ablation} presents ablation results examining consistency regularization, backbone architecture, and training strategy, each under the evaluation protocol noted in its row block. The consistency-regularization rows report external performance averaged over the held-out datasets for EfficientNet-B7 models trained on BUSI (a widely used public breast ultrasound benchmark) and BUS-BRA (the largest of our four datasets), confirming that the findings are not specific to a single training set. The backbone comparison uses internal five-fold cross-validation on both BUSI and BUS-BRA because sweeping seven encoders over all 12 cross-dataset pairs is computationally prohibitive. The training-strategy study uses the full-dataset configuration and reports the mean over all 12 cross-dataset pairs, testing whether the multi-task advantage persists beyond any single benchmark.

\begin{table*}[!t]
\centering
\caption{Ablation study examining consistency regularization, backbone architecture, and training strategy. The consistency-regularization rows report external performance averaged over the held-out datasets for BUSI- and BUS-BRA-trained EfficientNet-B7 models. For simplicity, the consistency-regularization ablation study does not use the chain augmentation applied in the final pipeline, so its absolute values are lower than, and not directly comparable to, the final results in Tables~\ref{tab:cross_dice} and~\ref{tab:cross_auc}. The backbone-comparison rows report internal five-fold cross-validation on BUSI and BUS-BRA. The training-strategy study evaluates single models trained using the full-dataset configuration and reports the mean over all 12 cross-dataset pairs (``All pairs''); under that block, DC and AUC each span two columns of the table grid because the scope no longer separates BUSI from BUS-BRA. The best value in each column is shown in \textbf{bold}.}
\label{tab:ablation}
\fontsize{7.5}{9}\selectfont
\setlength{\tabcolsep}{3pt}
\begin{tabular}{@{}>{\raggedright\arraybackslash}p{2.2cm}lcccc@{}}
\toprule
& & \multicolumn{2}{c}{\textbf{BUSI}} & \multicolumn{2}{c}{\textbf{BUS-BRA}} \\
\cmidrule(lr){3-4}\cmidrule(lr){5-6}
\textbf{Study} & \textbf{Configuration} & \textbf{DC} & \textbf{AUC} & \textbf{DC} & \textbf{AUC} \\
\midrule
\multirow{4}{*}{\parbox{2.2cm}{\raggedright Consistency reg.\\{\scriptsize external, B7}}} & Full ($\alpha{=}0.18$, $\lambda_{\text{mt}}{=}1.0$) & \textbf{.748} & \textbf{.820} & \textbf{.795} & \textbf{.868} \\
& Weak ($\alpha{=}0.10$) & .720 & .817 & .787 & .856 \\
& Strong ($\alpha{=}0.30$) & .698 & .803 & .781 & .840 \\
& No MTP ($\lambda_{\text{mt}}{=}0$) & .735 & .812 & .754 & .845 \\
\midrule
\multirow{7}{*}{\parbox{2.2cm}{\raggedright Backbone comp.\\{\scriptsize internal five-fold CV}}} & ResNet-50 (27.2M) & .796 & -- & .910 & -- \\
& ResNet-101 (46.2M) & .797 & -- & .911 & -- \\
& ResNeXt-50 (26.7M) & .799 & -- & .910 & -- \\
& SE-ResNet-50 (29.7M) & .801 & -- & .910 & -- \\
& EfficientNet-B3 (11.8M) & .802 & -- & .911 & -- \\
& EfficientNet-B5 (29.6M) & .806 & -- & .912 & -- \\
& EfficientNet-B7 (65.3M) & \textbf{.807} & -- & \textbf{.914} & -- \\
\midrule
\multicolumn{2}{@{}l}{\textit{Training strategy} (All pairs, B7, full-dataset)} & \multicolumn{2}{c}{\textbf{DC}} & \multicolumn{2}{c}{\textbf{AUC}} \\[1pt]
\multirow{7}{*}{} & Seg-only & \multicolumn{2}{c}{.740} & \multicolumn{2}{c}{--} \\
& Cls-only & \multicolumn{2}{c}{--} & \multicolumn{2}{c}{.791} \\
& Vanilla MT ($\alpha{=}0$) & \multicolumn{2}{c}{.735} & \multicolumn{2}{c}{.801} \\
& PCGrad~\citep{yu2020pcgrad} & \multicolumn{2}{c}{.746} & \multicolumn{2}{c}{.786} \\
& Uncertainty~\citep{kendall2018uncertainty} & \multicolumn{2}{c}{.755} & \multicolumn{2}{c}{\textbf{.828}} \\
& Ours, single model (w/o aug) & \multicolumn{2}{c}{.724} & \multicolumn{2}{c}{.813} \\
& Ours, single model & \multicolumn{2}{c}{\textbf{.764}} & \multicolumn{2}{c}{.818} \\
\bottomrule
\end{tabular}
\end{table*}

The ablation shows that consistency regularization is critical: both the weaker ($\alpha{=}0.10$) and stronger ($\alpha{=}0.30$) settings underperform the selected setting, and removing the MTP reduced the BUSI DC from 0.748 to 0.735. The backbone comparison confirms that the EfficientNet family is the strongest encoder on both training sets: EfficientNet-B5 and EfficientNet-B7 yield the best internal cross-validation DC values on BUSI (0.806 and 0.807) and BUS-BRA (0.912 and 0.914). On BUSI, the smaller EfficientNet-B3 (11.8M) also slightly outperforms the much larger ResNet-101 (46.2M), illustrating efficient compound scaling. On BUS-BRA, however, EfficientNet-B3 and the heavier ResNet-50/101, ResNeXt-50, and SE-ResNet-50 backbones are statistically indistinguishable (0.910--0.911, within the $\sim$0.006 fold standard deviation). Thus, the small-model advantage is specific to BUSI, whereas the EfficientNet-B5/B7 lead is replicated across both datasets. Our method outperforms both single-task specialists on their respective tasks and achieves the strongest DC among the multi-task configurations: segmentation DC increases from 0.740 to 0.764, and AUC increases from 0.791 to 0.818. Uncertainty weighting yields the highest AUC among the multi-task configurations (0.828). Chain augmentation provides a 0.040 increase in DC (0.724$\rightarrow$0.764). The BUS-BRA consistency-regularization ablation reproduces the BUSI trends on a second training set: the selected configuration is best (DC $=$ 0.795), both weaker ($\alpha{=}0.10$, 0.787) and stronger ($\alpha{=}0.30$, 0.781) consistency underperform it, and removing the minimal-tumor penalty lowers DC from 0.795 to 0.754. This confirms that the full regularized formulation generalizes beyond a single training set.

\FloatBarrier
\subsection{Computational Cost}
\label{sec:compute}

To address the practical cost of the proposed model relative to lighter alternatives, Table~\ref{tab:compute} reports model size and runtime. All measurements use $256\times256$ input on a single NVIDIA Tesla V100 (SXM2, 32~GB) at batch size~16; training-step time is one optimizer step (forward + backward + update), and inference is single-image latency. Our EfficientNet-B7 + DeepLabV3+ multi-task model has 65.3M parameters, with a single-image inference latency of 58.6\,ms (17.1\,img/s). Compared with a lighter single-task EfficientNet-B0 alternative (4.9M parameters, 9.9\,ms inference), our model has approximately 13 times as many parameters and is 6 times slower at inference. This additional capacity is associated with the model's cross-dataset performance and supports the BI-RADS consistency mechanism. The single-task EfficientNet-B7 segmentation baseline runs at 30.6\,ms/image; the additional inference cost of the multi-task model over this baseline (about 28\,ms) corresponds to a redundant second encoder forward pass in the measured PyTorch implementation, which recomputes the shared encoder features for the classification branch. Caching the shared encoder activations at inference would eliminate this overhead without changing the trained model, so the reported multi-task latency is conservative.

\begin{table*}[t]
\centering
\caption{Computational cost on NVIDIA Tesla V100 (SXM2, 32~GB), $256\times256$ inputs. Parameter counts are exact; runtimes are wall-clock measurements. ``Train s/step'' is one optimizer step at batch size~16; ``Infer ms/img'' is single-image inference latency, and img/s is its reciprocal ($1000/\text{ms}$).}
\label{tab:compute}
\fontsize{9}{11}\selectfont
\setlength{\tabcolsep}{8pt}
\begin{tabular}{@{}lcccc@{}}
\toprule
\textbf{Model} & \textbf{Params (M)} & \textbf{Train (s/step)} & \textbf{Infer (ms/img)} & \textbf{img/s} \\
\midrule
Ours (EfficientNet-B7, MT)      & 65.3 & 0.509 & 58.6 & 17.1 \\
Single-task seg (EfficientNet-B7) & 65.1 & 0.390 & 30.6 & 32.7 \\
Lighter alt.\ (EfficientNet-B3) & 11.7 & 0.112 & 15.3 & 65.4 \\
Lighter alt.\ (EfficientNet-B0) &  4.9 & 0.059 &  9.9 & 101.0 \\
\bottomrule
\end{tabular}
\end{table*}

\section{Discussion}
\label{sec:discussion}

This work has three central findings. First, despite not using foundation-model pretraining, our conventional EfficientNet-B7 multi-task model is competitive with recent SAM-based methods in cross-dataset breast ultrasound segmentation. Across all 12 train--test pairs in our complete directed evaluation of four public datasets, the model reaches a mean external DC of 0.786, including 0.826 on BUSI$\rightarrow$UDIAT. It also predicts malignancy, which the segmentation-only SAM adaptations do not. Unlike earlier external MTL reports limited to one cohort, one direction, or two-domain 3D transfer, this design exposes the model to every ordered source--target combination and therefore tests whether its advantage persists across the full pattern of asymmetric dataset shifts. Second, incorporating BI-RADS morphological descriptors improves both tasks. It raises DC from 0.740 to 0.764 and AUC from 0.791 to 0.818 relative to single-task specialists, and it achieves the highest DC among the task-agnostic multi-task remedies while remaining competitive on AUC. The gain therefore reflects a clinically grounded link between the classification and segmentation tasks rather than generic gradient de-confliction. Third, the learned morphology weights follow the same ordering on all four datasets: Roughness $>$ Texture $>$ Area $>$ non-compactness $(1-C)$. This ordering persists despite substantial domain shift across the four datasets, including differences in scanner equipment, image size and resolution, lesion size, and lesion area relative to the image (Tables~\ref{tab:datasets} and~\ref{tab:size_strat}). The ordering is consistent with BI-RADS clinical criteria and suggests that the morphology-to-malignancy relationship is stable across domains. We examine these findings below, together with the boundary-distance behavior and a failure analysis that exposes a concrete cause of poor transfer.

The boundary-distance results (Table~\ref{tab:cross_hd95}) reinforce the findings suggested by the DC values. Models trained on UDIAT remain the weakest source models according to both metrics: they have the lowest mean DC (0.709) and the highest mean of the pairwise median HD95 values (14.93\,px). BUSI is the hardest target: averaged over sources, its single-model median HD95 is 20.6\,px (Table~\ref{tab:cross_hd95}, bottom row). UDIAT is the easiest target, at 4.8\,px, because its small lesions have short boundaries. UDIAT$\rightarrow$BUSI is the worst DC cell (0.675). Its overall median HD95 is only 20.0\,px because most BUSI test lesions are small or medium, so the median is not dominated by the large-lesion failures. The error is concentrated in large lesions, for which the tercile median HD95 reaches 46.1\,px (Table~\ref{tab:size_strat}); Figure~\ref{fig:seg_failure} shows representative examples. The two metrics identify different best-performing source datasets. Models trained on BrEaST yield the best DC (0.792), whereas models trained on BUSI yield the lowest median HD95 (8.97\,px). Thus, BrEaST yields more complete overlap, whereas BUSI yields tighter boundaries. The ensemble lowers HD95 in 11 of 12 cells, from 12.2 to 9.3\,px on average. The largest gains occur for transfers into BUSI.

Beyond the cross-dataset matrix, the ablation in Table~\ref{tab:ablation} isolates the contribution of the multi-task formulation. Averaged over all 12 cross-dataset pairs, the full single-model configuration improves segmentation relative to a dedicated single-task segmentation network (DC, 0.764 vs.\ 0.740) and malignancy classification relative to a dedicated single-task classifier (AUC, 0.818 vs.\ 0.791). Thus, the model does not improve one task at the expense of the other, a central risk of naive multi-task training. The source of this gain is informative. A vanilla shared-encoder model ($\alpha{=}0$) does not outperform the single-task segmentation baseline, indicating that destructive task interference is present in this setting. The task-agnostic remedies that we evaluated, gradient surgery and uncertainty weighting, reduce but do not eliminate this interference. Only when the classification head is linked to morphological descriptors computed from the segmentation output does the model surpass both single-task specialists. Removing the auxiliary minimal-tumor penalty (No MTP) lowers DC on both training sets, showing that this term stabilizes segmentation, whereas the morphology-classification consistency term remains the primary bridge between the two tasks. We interpret these results as evidence that the consistency term supplies a clinically grounded link between the tasks rather than a purely optimization-level de-confliction: the two heads are encouraged to agree on features, including margin irregularity, area, and texture, that are themselves predictive of malignancy. The stability of the learned weight ordering across domains (Roughness $>$ Texture $>$ Area $>$ Compactness across all four datasets) is consistent with this interpretation. A coupling anchored in descriptors whose relative importance is stable across centers may be less prone to overfitting the idiosyncrasies of any single source.

These results also elucidate how a conventional architecture compares with the foundation-model methods that increasingly dominate domain-generalized ultrasound segmentation. On the three external target datasets evaluated using BUSI as the training source, our ensemble model (mean DC, 0.805) is on par with the publicly released, BUSI-trained Nora model (mean DC, 0.806). Our reproduction of Nora's BUSI$\rightarrow$UDIAT result (0.839) also matches its published value, lending confidence to the comparison. An ImageNet-pretrained EfficientNet-B7, trained on several hundred to several thousand breast ultrasound images, reaches performance comparable to that of a SAM backbone. It also predicts malignancy, which the segmentation-only SAM adaptations do not. These results suggest that foundation-model pretraining may contribute less on this task than single-direction comparisons imply. The unified bidirectional cross-dataset protocol makes this result visible: across the full 12-pair matrix, our method outperforms the reimplemented USE-MiT (mean DC, 0.786 vs.\ 0.719), and HA-Net underperforms on every BUSI-source target (Table~\ref{tab:sota}). Published single-direction figures obtained under each method's own protocol are therefore best interpreted as reference points rather than matched comparisons. Re-evaluating relevant baselines under one protocol is arguably a prerequisite for credible cross-dataset claims.

Finally, the failure analysis identifies a concrete and actionable cause of poor transfer. In the worst direction, UDIAT$\rightarrow$BUSI, the model does not merely segment imprecisely; it misses the lesion entirely and activates on a fixed corner region (Figure~\ref{fig:seg_failure}) that coincides with a dataset-specific annotation marker present in BUSI but absent from the UDIAT training images. The model appears to rely on this spurious cue because it was absent from the UDIAT training data. This is a dataset-specific spurious correlation rather than a generic segmentation difficulty, and it compounds the tumor-size mismatch quantified above. Our findings align with evidence that performance in ultrasound analysis depends not only on the model and pipeline but also on data quality~\citep{vallez2025busuclm}. This finding also reinforces the need for external, multidirectional evaluation: an internal split could not have exposed this failure because the spurious cue is consistent within each dataset and becomes harmful only across datasets.

\section{Limitations}
\label{sec:limitations}

This study has several limitations. First, the multi-task formulation substantially improves segmentation, whereas its benefit to malignancy classification under domain shift is smaller. We therefore present malignancy classification as a secondary, auxiliary outcome whose primary role is to supply the morphological consistency signal during training rather than to serve as a standalone diagnostic tool. External malignancy AUC is more variable than external segmentation DC, particularly with UDIAT as the source dataset (Table~\ref{tab:cross_auc}), and even a substantially higher AUC would not by itself establish clinical deployability. Second, all images are resized to $256\times256$, which is coarse relative to their native resolutions (Table~\ref{tab:datasets}). Consequently, HD95 is computed on resized masks, and we do not characterize the effect of resolution on the boundary-sensitive morphological features at the core of our method, including roughness and compactness. Third, the consistency strength $\alpha$ is tuned for each dataset rather than set using a single rule, which complicates application to a new cohort without retuning. Although $\lambda_{\text{mt}}$ is fixed at 1.0 in our experiments, its optimal value may also vary across data distributions. Finally, our evaluation includes only four international datasets. Although this is, to our knowledge, among the broadest external evaluations reported for breast ultrasound segmentation, the number of eligible datasets is limited because our multi-task formulation requires both pixel-level segmentation masks and image-level malignancy labels. Relatively few public breast ultrasound datasets provide both annotation types. Additional datasets with both annotation types, together with prospective multi-reader studies, would further strengthen claims of generalizability.

\section{Conclusion}
\label{sec:conclusion}

We introduced a novel, clinically grounded task-coupling mechanism that converts BI-RADS-inspired features from a soft segmentation mask into a differentiable consistency constraint on malignancy classification. To our knowledge, this is the first breast ultrasound method to make morphology derived from its own soft mask the explicit end-to-end bridge between segmentation and malignancy classification. We also provide the first complete directed external evaluation of a joint 2D B-mode model across four independent datasets, covering all 12 source--target pairs, together with a matched all-pairs comparison against single-task specialists, naive MTL, PCGrad, and uncertainty weighting. In these matched comparisons, the proposed single-model configuration improves both segmentation (mean DC, 0.764 vs.\ 0.740) and classification (mean AUC, 0.818 vs.\ 0.791) over dedicated single-task models. Using an EfficientNet-B7 backbone, the ensemble achieves a mean external DC of 0.786, including 0.826 on BUSI$\rightarrow$UDIAT, and remains competitive with published SAM-based domain-generalization results without foundation-model pretraining while additionally predicting malignancy. The identical learned ordering of morphological weights across four international datasets provides further evidence that this coupling captures a stable and interpretable relationship rather than a dataset-specific optimization artifact. Together, these findings establish both a new clinically grounded MTL mechanism and a substantially broader standard of external evidence for joint breast ultrasound analysis. Future work will investigate domain adaptation, more robust cross-dataset malignancy classification, and evaluation on additional datasets. Code and trained models will be made publicly available upon acceptance.

\FloatBarrier

\section*{CRediT authorship contribution statement}
\textbf{Jingru Zhang:} Conceptualization, Methodology, Software, Validation, Formal analysis, Investigation, Data curation, Visualization, Writing -- original draft. \textbf{Saed Moradi:} Conceptualization, Methodology, Supervision, Validation, Investigation, Writing -- review \& editing. \textbf{Ashirbani Saha:} Conceptualization, Methodology, Supervision, Funding acquisition, Resources, Project administration, Investigation, Writing -- review \& editing.


\section*{Acknowledgments}
This study was supported in part by the NSERC Discovery Grant. The computing resources used in this work were provided by PHRI and CREATE.


\end{document}